\documentclass[sigconf, nonacm]{acmart}

\AtBeginDocument{%
  }

\begin{document}

\title{Multi-Point Proximity Encoding For Vector-Mode Geospatial Machine Learning}

\author{John Collins}
\orcid{0009-0005-6490-2913}
\affiliation{
	\institution{Odyssey Geospatial LLC}
	\city{Brentwood}
	\state{New Hampshire}
	\country{USA}
}
\email{john@odyssey-geospatial.com}

\begin{abstract}
Vector-mode geospatial data -- points, lines, and polygons -- 
must be encoded into an appropriate form in order to be used with traditional
machine learning and artificial intelligence models. 
Encoding methods attempt to represent a given shape as a vector 
that captures its essential geometric properties. 
This paper presents an encoding method 
based on scaled distances from a shape to a set of
reference points within a region of interest.
The method -- Multi-Point Proximity (MPP) encoding --
can be applied to any type of shape, enabling the parameterization
of machine learning models with encoded representations 
of vector-mode geospatial features.
We show that MPP encoding possesses the desirable properties 
of shape-centricity and continuity,
can be used to differentiate spatial objects
based on their geometric features, 
and can capture pairwise spatial relationships with high precision.
In all cases, MPP encoding is shown to perform better than an alternative
method based on rasterization. 
\end{abstract}

\maketitle

\section{Introduction}

The distinction between vector-mode and raser-mode approaches to spatial analysis
has existed since the dawn of Geographical Information Systems (GIS).
In raster mode, data are represented as values on a regular grid. 
The grid itself is referenced to a coordinate system,
and the spatial properties of each data element is implied by its position in the grid. 
Vector-mode approaches instead represent geospatial features as 
discrete points, lines, and polygons, 
with data values taking the form of attributes assigned to such objects.
Both modes are widely used, and have unique strengths and limitations.
Any image from a satellite or airborne sensor
is naturally represented as a raster, for example.
Discrete entities such as road segments, building footprints,
and administrative boundaries are most typically defined as vector-mode data.

In recent years there has been great interest in applying 
the methods of Machine Learning (ML) and Artificial Intelligence (AI)
to geospatial problems. 
Within this area, there is a heavy emphasis on what are essentially 
raster-mode methods. This is hardly surprising, since 
so many advances in ML have been in the area of image processing --
including image classification \cite{Krizhevsky17}   
and object detection \cite{Redmon16}. 
Such methods 
find natural application in the processing of remotely sensed data, for example
in classifying land cover types \cite{Gavade25}\cite{Ma17}, 
characterizing zones of growth and development \cite{Mushore17},
and distinguishing features in the landscape \cite{Kotaradis21}\cite{Zhang23}. 

There is great potential for applying ML and AI approaches 
to vector-mode geospatial analysis problems. But such methods contend with 
the fact that most traditional machine learning algorithms -- 
classifiers, regressors, neural networks and so on -- 
are not designed to ingest vector data 
in the formats in which they are commonly represented. 
The most common class of vector data representations are those derived from 
the Open Geospatial Consortium's (OGC) Simple Feature Architecture \cite{ogc:sfa}.
This specification is embodied by a number of specific formats, including 
Well-Known Text (WKT) \cite{ogc:wkt}, GeoJSON \cite{rfc7946}, and others.
Under these formats, LineString and Polygon objects are represented
as arbitrary-length lists of coordinate pairs defining linear paths
and polygon boundaries repectively. Their irregular format makes them difficult to 
use in ML models, which typically require inputs to be
consistently sized vectors.

To address this difficulty, the field of Spatial Representation Learning (SRL) 
has gained prominence in recent years \cite{Mai24} \cite{Wozniak21}.
The goal of SRL is to 
represent geospatial objects as vectors that capture their spatial properties,
which can then be used in further ML/AI processing. 
The topic is recognized as a critical component enabling further evolution 
of geospatial AI \cite{Mai25}. 
While a number of solutions have been proposed,
many have been developed for specific types of geometries -- for example
methods to endode only points or only polygons. 
This specificity partly hinders the application
of SRL developments to general geospatial analysis problems,
to the extent that \cite{Mai25} specifically 
highlights the need for 
"a unified representation learning model 
that can seamlessly handle different spatial data formats".

This paper proposes a method that partially address this need for general-purpose encoding
of geometric objects. A straightforward adaptation of an existing point encoding approach
\cite{Yin19}, allows its application to shapes of any type. 
Specifically using an approach that we call Multi-Point Proximity (MPP) encoding,
one can represent any shape based on its distance to a number of reference points
within a region of interest. Re-scaling these distances and arranging them as a vector
yields a convenient ML-compatible representation.

Furthermore we present a framework for assessing the degree to which
any such encoding captures measurable properties of the encoded shape.
The framework consists of training models to estimate measurable geospatial properties 
directly form encoded representations. 
Performance of such models 
is a proxy for the degree to which the encodings capture the properties.
Using this framework, we demonstrate that MPP encoding
performs better than an a specific alternative based on rasterization. 

The remainder of this paper is organized as follows.
Section \ref{SecEncodingsAndEmbeddings}
discusses the general background and recent developments in SRL,
and outlines some of the key differences among SRL methods and applications.
Sections \ref{SecMPP} and \ref{SecDIV}
respectively present MPP encoding and an alternative baseline approach 
based on Discrete Indicator Vectors (DIV).
Section \ref{SecShapeDiff} illustrates how general-purpose encoding methods
support differentiation of spatial features based on their geometric properties.
Section \ref{SecProperties} shows how MPP and DIV perform relative to various criteria 
for encodings, and shows the degree to which they capture properties of 
individual shapes as well as pairwise relationships between geospatial features.
We conclude by discussing possible applications and directions for further work.

\section{Encodings and Embeddings}
\label{SecEncodingsAndEmbeddings}

According to the OGC model, a spatial \textit{feature} $\mathbf{f}$
consists of a \textit{geometry} $\mathbf{g}$
and a set of \textit{attribute values} $\mathbf{v}$.
\begin{displaymath}
	\mathbf{f} = (\mathbf{g}, \mathbf{v})
\end{displaymath}
The geometry $\mathbf{g}$ consists of the feature's representation as a 
Point, LineString, Polygon, or some collection thereof. 
The attributes $\mathbf{v}$ consist of any qualitative or quantitative properties
of the feature.
Vector-mode spatial representations are not directly ML-compatible,
and require \textit{geometric encoding} in order to be used in SRL. 
\begin{equation}
	\mathbf{g}^* = \mathrm{ENC}^\mathbf{g}(\mathbf{g})
\end{equation}
where $\mathrm{ENC}^\mathbf{g}$ is a function that maps a geometry 
into a $d_g$-dimensional space
in such a way as to capture its geometric properties. 
The analogous operation of \textit{attribute encoding}
\begin{equation}
	\mathbf{v}^* = \mathrm{ENC}^\mathbf{v}(\mathbf{v})
\end{equation}
creates vector representations of any attribute that is not naturally 
a numerical quantity.
Both geometric and attribute encoding are most often algorithmic processes; they 
are effected via a well defined processing sequence or closed-form transformations. 
Examples include "rasterization" of vector-mode geometries, and one-hot encoding for 
categorical attributes.

The central problem of SRL
is to find an \textit{embedding} $\mathbf{m}$
for either a feature $\mathbf{f}$
or a set of $N$ features $\{\mathbf{f}_i; i \in 1..N\}$.
\begin{equation}
	\label{EqEmbeddingDef}
	\mathbf{m}=\mathrm{EMB}(\{
	   ( \mathbf{g}_i^*, \mathbf{v}_i^*) :
	     i \in 1..N
	\}),
\end{equation}
That is, $\mathrm{EMB}$ maps encoded geometries and attributes
into a space of dimension $d_m$ to enable further ML / AI processing. 
In contrast to encodings, embeddings typically are \textit{learned} from training examples,
often  using a self-supervised objective.
In the geospatial context, a useful choice is a triplet loss function \cite{Hoffer15}, 
where the model is trained to yield similar embeddings for locations 
that are close to one another,
and divergent embeddings for features that are more distant 
\cite{Jean19}\cite{Sentiance}\cite{Wang20}.
This is often described as leveraging the "first law of geography", 
that things closer together
tend to be more similar than things further apart \cite{Tobler70}. 

With these definitions, geometric encoding can be seen as a separable component
of the geospatial embedding process. 
Consideration of methods and algorithms for representing shapes
can proceed independently of discussions about algorithms and processes for embeddings.
Furthermore as a necessary initial step in the embedding process, 
the key criterion for a good geometric encoding is minimal information loss: 
it should faithfully capture all relevant spatial properties 
of the geometries that it encodes. 

Existing works on location encoding were recently 
reviewed and categorized in \cite{Mai22},
and the model implied by equations 1-3 above
encompass much of their taxonomy.
The review primarily focuses on encodings for Point geometries.
The authors distinguish \textit{single-point} and \textit{aggregate} methods,
which respectively correspond to sets of size $1$ and $>1$ in Equation \ref{EqEmbeddingDef}.
Methods for Point encoding include rasterization,
in which a study region is divided into non-overlapping tiles,
and a Point is encoded as a one-hot vector indicating which tile it
falls into \cite{Tang15}. 
Alternate approaches involve encoding a Point's $(x,y)$ components
using sinusoidal functions \cite{Mai20b} inspired by those used for positional encodings
in transformer models \cite{Vaswani17}.
Aggregate approaches to Point encoding include computing a Point's scaled distance
to a set of reference points within a Region Of Interest (ROI) \cite{Yin19},
and methods that train a Neural Network considering all points within 
a neighborhood \cite{Qi17}.

Various methods have been proposed for handling other types of geometries. 
A noted approach to encoding LineStrings 
is to apply a sequential neural network 
to the ordered encodings of the points that compose it \cite{Xu18}. 
For encoding of Polygons, the authors note prior work based on 
encodings of points sampled from within a polygon's bounding box \cite{Mai20a}.
They describe the method as a "first step" towards general
Polygon encoding, and state that more work needs to be done in this area.

More recently \cite{Cui25} proposed a method for encoding Lines and Polygons
based on one-hot vectors indicating proximity to refence points
in the immediate vicinity of the shape to be encoded. 
A representation is learned by training a transformer-based model 
using self-supervised tasks predicting perturbations to the encoded shapes.

In spite of all recent advances, there is yet no widely accepted
approach for geometric encoding that can handle different types of
geometries using a consistent approach. The need for such a method
was reiterated in 
a more recent work describing a vision for next generation of 
geospatial ML and AI \cite{Mai25}.
Next we present two approaches that have some potential to fill this need.

\subsection{Multi-Point Proximity Encoding}
\label{SecMPP}

Among the works catalogued in \cite{Mai22} 
is a kernel-based method for encoding GPS point locations globally,
originally presented by \cite{Yin19}, under the name GPS2Vec.
It starts by defining grids of reference points relative to zones of 
the Universal Transverse Mercator (UTM) coordinate system. 
To encode a Point, GPS2Vec computes its distance to the reference points,
and applies a negative exponential kernel function to scale the distance value. 
These scaled distances make up the encoding.

There is nothing about the method that limits its use to UTM zones;
it works just as well in any rectangular coordinate system.
Within any ROI, one can specify an ordered set of $N$ reference points
$\mathbf{r} = [r_i: i \in 1..N]$.
Then to define the distance $\mathrm{dist}(\mathbf{g}, r_i)$
between a point $r_i$ and an arbitrary type of geometry $\mathbf{g}$, 
we can follow the standard practice 
of taking the minimum distance to any point on or in $\mathbf{g}$.
\begin{equation}
	\mathrm{dist}(\mathbf{g}, r_i) = \mathrm{min}{||p-r_i||: p \in \mathbf{g}}
\end{equation}
Given this definition, the logic of GPS2Vec can be applied for any type of geometry. 
\begin{equation}
	\label{MppDefEq}
	\mathbf{g}^*_\mathrm{MPP} = 
	\mathrm{MPP}(\mathbf{g}) = 
	[\mathrm{exp}(-\mathrm{dist}(\mathbf{g}, r_i) / s) : r_i \in \mathbf{r}],
\end{equation}
where $s$ is a scaling factor. 

\begin{figure}
	\centering
	\includegraphics[width=\linewidth]{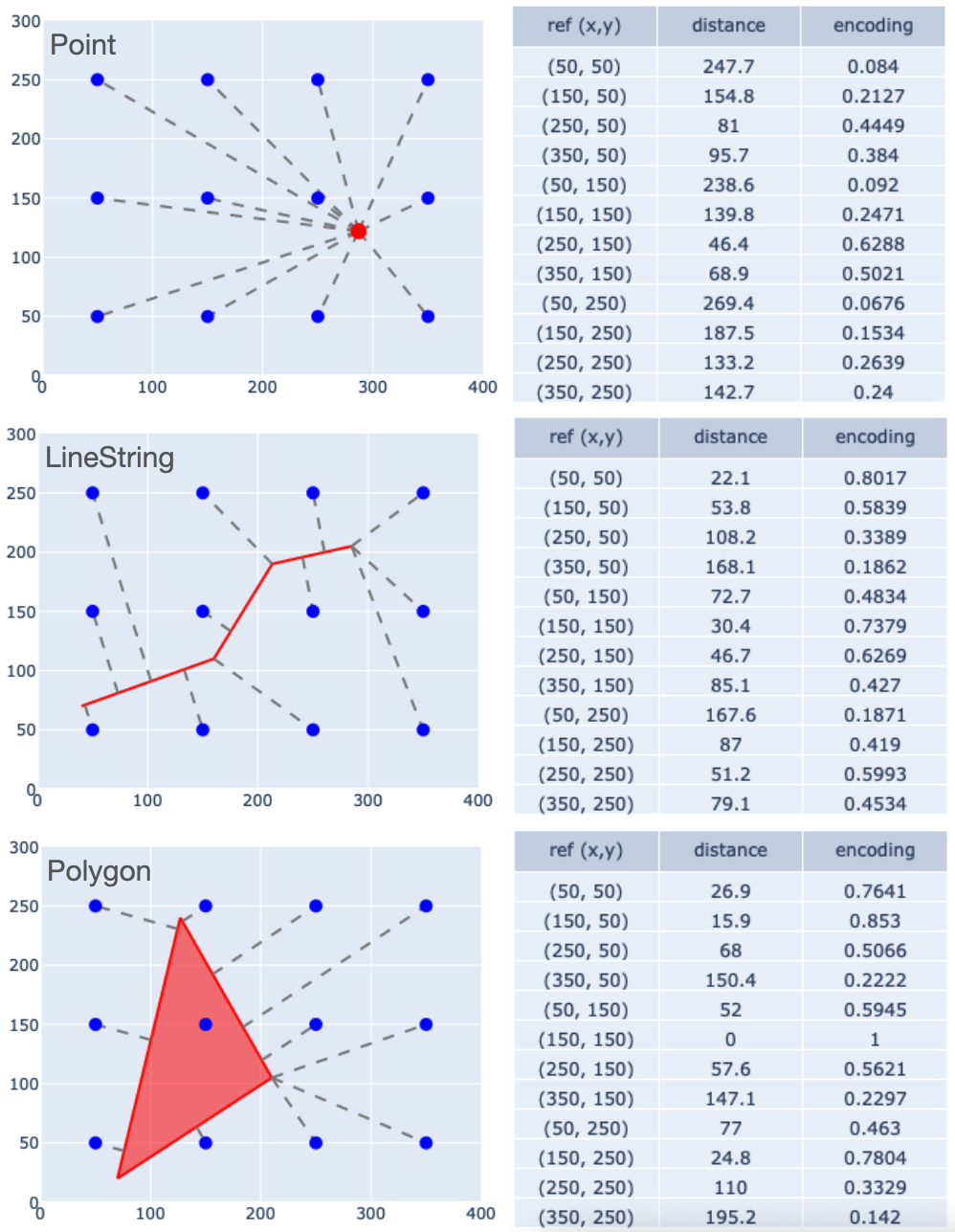}
	\caption{Multi-Point Proximity (MPP) encodings for a Point, LineString, and Polygon.
		Blue dots represent reference points within the 400-by-300 region,
		and gray dashed lines indicate their closest points used for
		distance calculations.}
	\label{MppExampleFig}
\end{figure}

Figure\ref{MppExampleFig} is an example of MPP encodings for three different types of geometries.
The ROI measures $400 \times 300$ units, and we have defined a $3 \times 4$ grid of reference points
with a spacing of 100 units. The scaling factor $s$ from Equation (\ref{MppDefEq}) 
is defined to be 100 -- the same as the
reference point spacing. 
This yields conformable representations of all three geometry types.
Note that the distance for the reference point inside the Polygon is zero, 
and the associated element
of the encoding is scaled to 1.0 via the kernel function. 
Since the negative exponential kernel 
is monotonically decreasing and positive, all elements of the encoding 
lie in the interval $(0, 1]$,
which is a convenient range for use in ML and AI models.

\subsection{Discrete Indicator Vector Encoding}
\label{SecDIV}

Any vector-format geometry can be converted to a raster representation. 
One of the earliest attempts at integrating geospatial information
with ML/AI models \cite{Tang15} incorporates raster map data 
for various features, including administrative boundaries,
which are typically derived from vector-format Polygons.
Another approach called Loc2Vec \cite{Sentiance} incorporates vector-mode data from
OpenStreetMap by rendering different feature types into small raster tiles,
using a particular set of styling rules. Both of the latter approaches 
incorporate the rasters into downstream models 
using Convolutional Neural Network (CNN) processing.

Another example of rasterization, which is even more general than the 
latter two approaches, was mentioned above: 
one-hot encoding for a tile that contains a given Point.
It is straightforward to define the encoding in a way that
apples to an arbitrary geometry $\mathbf{g}$. 
Define an ordered set of $N$ non-overlapping tiles 
$\mathbf{t} = [t_i: i \in 1..N]$ that collectively cover a ROI. 
One can define a Discrete Indicator Vector (DIV)
indicating which tiles intersect $\mathbf{g}$.
\begin{equation}
	\mathbf{g}^*_\mathrm{DIV} = \mathrm{DIV}(\mathbf{g}) =
	[0 \thickspace \mathrm{if} \mathbf{g} \cap t_i \thickspace = \emptyset 
	\thickspace \mathrm{else} \thickspace 1: t_i \in \mathbf{t}]
\end{equation}
So $\mathbf{g}^*_\mathrm{DIV}$ is a $n$-element vector encoding $\mathbf{g}$.
As with any rasterization, the level of fidelity 
depends on the resolution -- i.e. the size and number of tiles. 

\begin{figure}[b]
	\centering
	\includegraphics[width=\linewidth]{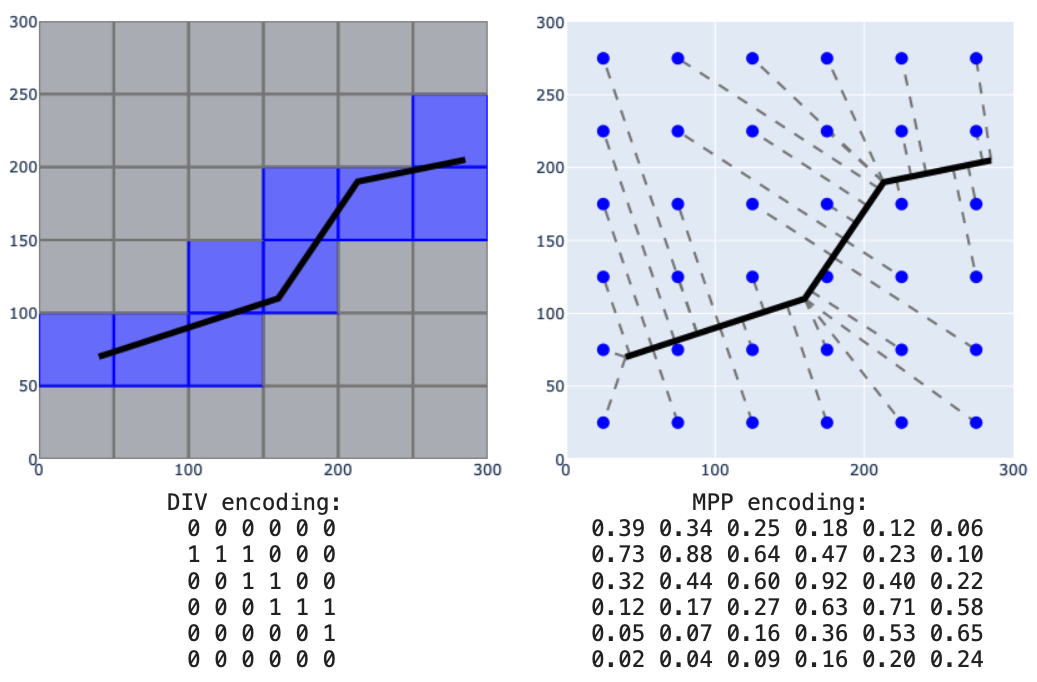}
	\caption{DIV and MPP encodings for a LineString geometry.}
	\label{CmpMppDivFig}
\end{figure}

Figure \ref{CmpMppDivFig} illustrates the idea in comparison to MPP encoding. 
We have defined a single LineString within a $300 \times 300$ region.
TheDIV encoding is based on tiles measuring $50 \times 50$ units.
The MPP encoding uses the centroids of those tiles as reference points, so both methods
yield a 36-element encoding, and could be said to have the same "resolution".

It is easy to see that any LineString -- and indeed any Polygon -- 
intersecting the same set of tiles
would have the exact same DIV encoding. 
In contrast, MPP yields a more continuous representation
for which slight shifts in the LineString coordinates would yield corresponding changes 
in the encoding.
So clearly the MPP approach is the more sensitive for encoding geometric properties.
As the DIV aproach corresponds more closely to previously proposed methods, it will be
used as a baseline for comparison in the following sections.

The DIV encoding method bears some similarity to 
the serialization approach proposed in \cite{Cui25}, 
which uses one-hot indicator vectors to encode nearest reference points. 
But the reference points are defined relative to the centroid of each geometry
rather than relative to a larger frame of reference as proposed above. 
The approach produces a set of such indicator vectors corresponding to 
the vertices of the input geometry, which are then used
to create a learned embedding via a transformer-based model.

\subsection{Shape Differentiation From Geometric Encodings}
\label{SecShapeDiff}

\begin{figure}
	\centering
	\includegraphics[width=\linewidth]{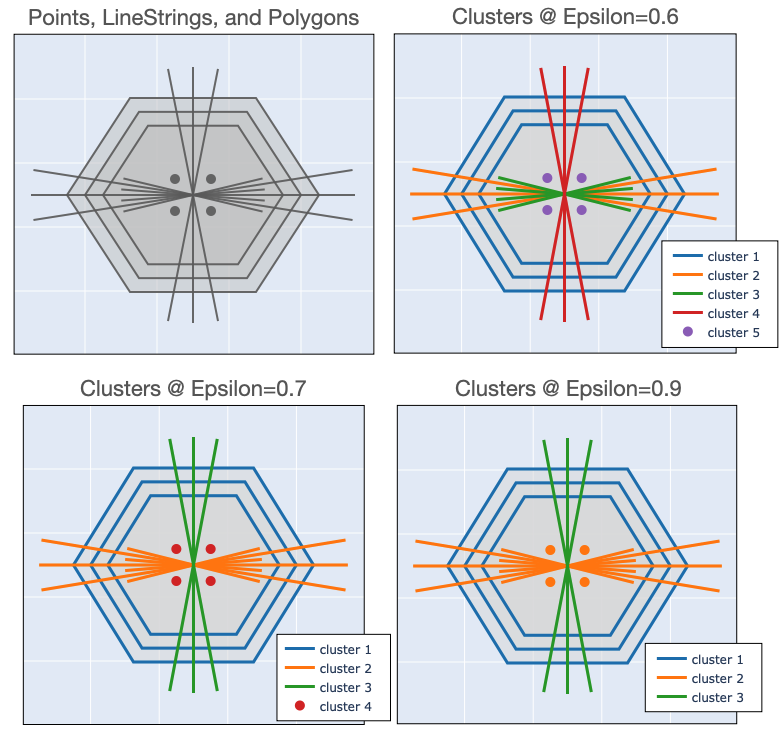}
	\caption{Clusters of Polygons, LineStrings, and Points based on geometric encodings.}
	\label{FigShapeClusters}
\end{figure}

To see the potential utility
of general-purpose geometric encoding, consider Figure \ref{FigShapeClusters}.
It shows some shapes of various types: three hexagonal Polygons,
four Points, and three groupings of LineStrings differentiated by their general orientation
and length.
For this example we defined a $5 \times 5$ grid of reference points
covering the square region, and used an MPP encoder to represent
each shape as a 25-element vector.

Rather than examining the vectors directly, it is illustrative to show what happens when they 
are used as inputs to a DBSCAN clustering algorithm. This looks for clusters
of shapes that lie near one another in the 25-dimensional encoding space 
rather than in the 2-dimensional reference frame. 

DBSCAN uses a parameter $\epsilon$ (epsilon) as a distance threshold 
for deciding whether a pair of points
belong to the same cluster. 
At a fairly low value of $\epsilon = 0.6$, DBSCAN finds five clusters, corresponding to 
recognizable categories of shape type, orientation, and length. 
The three hexagonal Polygons form a cluster, the four Points another cluster, and so on. 
At a slightly larger $\epsilon$, the two groups of long and short nearly-horizontal LineStrings 
merge into a single cluster, but are still distinguished from the vertical LineStrings, 
the Points, and the Polygons.
At an even larger $\epsilon$ the Point objects get merged with the horizontal LineStrings.

This simple example shows that while the vector-format definitions of these geometries 
are difficult to use in ML models, their geometric encodings have no such limitations. 
The general-purpose nature of the MPP approach 
alows us to handle different types of shapes consistently.
And although these shapes occupy overlap one another within the reference frame, 
and most have exactly the same centroid, 
ML processing clusters them by leveraging
similarities and differences along dimensions
of orientation, length, and form. 
It is even possible for different types of shapes to cluster together 
based on their proximity in encoding space. 

\section{Properties Of Geometric Encodings}
\label{SecProperties}

Having defined a pair of general-purpose encoding methods, this section
describes and compares the fidelity and utility of the representations. 

\subsection{Encodings as Approximations}

Both of the encoding methods described above are approximate, in that they do not
in general give an exact representation of the shape. 
This is easily understood when considering the DIV approach for grid cells: 
the location of a point, the position of a line segment, or the location of a polygon border
are only known to the resolution of the grid cells used in the encoding.

\begin{figure}[h]
	\centering
	\includegraphics[width=\linewidth]{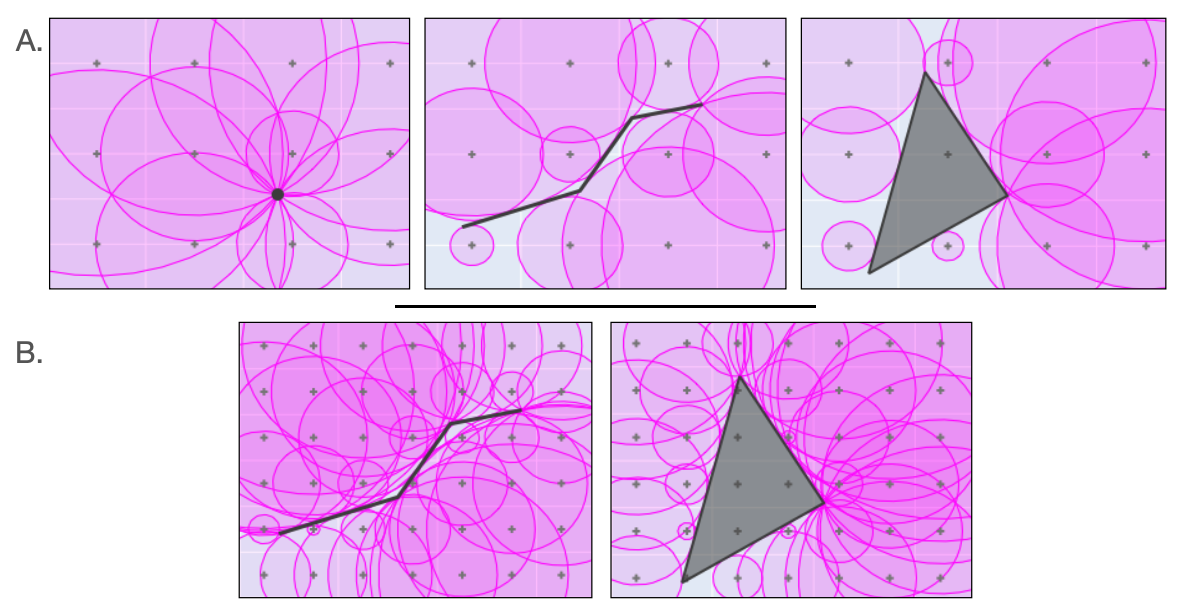}
	\caption{
		A. Exclusion zones around reference points for an encoded Point, LineString, and Polygon.
		Pink circles represent information that can be extracted from the encoding itself,
		providing hints about the form of each encoded geometry.
	    B. Improved representation for the LineString and Polygon geometries using a finer resolution.
	}
	\label{DecodingFig}
\end{figure}

Analogously, MPP encodings are limited by resolution, where "resolution" is understood to 
reflect the spacing between reference points.
Because the elements of an MPP encoding
are a monotonic function of the minimum distance to reference points,
one can compute that distance by inverting Equation \ref{MppDefEq}. 
If $g_i$ is an element of an encoding $\mathbf{g}^*_\mathrm{MPP}$,
then the distance to the associated reference point $r_i$ is
\begin{displaymath}
	d_i = -\ln g_i / s.
\end{displaymath}
This tells us two useful things about the geometry $\mathbf{g}$:  
\begin{itemize}
	\item One point of $\mathbf{g}$ lies a distance $d_i$ from the point $r_i$,
	\item No part of $\mathbf{g}$ lies any closer to $r_i$.
\end{itemize}
That is, each element of $\mathbf{g}^*_\mathrm{MPP}$ defines a kind of "exclusion zone"
around reference point $r_i$ to which the geometry $\mathbf{g}$ is tangent.
To see how this relates to the fidelity of the encoding, consider Figure \ref{DecodingFig}A,
which shows these exclusion zones for a Point, a LineString, and a Polygon
relative to a grid of reference points.
For the Point geometry, all exclusion zones intersect at the encoded point; in this case 
the geometry is encoded exactly and could be determined analytically.
For the LineString and Polygon, the encodings give a general idea about the form of the original shape,
but leave room for ambiguity. Figure \ref{DecodingFig}B shows that this ambiguity is reduced
when using a finer resolution -- i.e. decreasing the spacing between the reference points. 

\subsection{Shape-Centricity and Continuity}

Some proposed methods for encoding LineStrings and Polygons
process the list of coordinates that define them, for example \cite{Xu18}. 
Such an approach has a drawback: 
depending on the details of the encoding method,
the order in which the coordinates are processed may affect the results.
There would be a strong potential for such an outcome for example
if processing encodings for a list of polygon vertices using a recurrent neural network.

For this reason, a desirable property of any geometric encoding is \textit{shape-centricity}:
the encoding should only depend on the form of the geometry itself, and not on the
specifics of how its vertices are specified \cite{Mai24}.
For example, the border of a polygon consists of a linear ring of coordinates, forming a loop.
An encoding should not vary depending on which of the vertices is chosen as the begining of the loop.
Such an encoding is said to have  
\textit{vertex loop transformation invariance}. 
Another possible problem case consists of the presence of vertices whose omission would not change the form of a geometry,
for example a vertex at the midpoint of a straight line segment. 
An encoding that is unaffected by such vertices is said to have \textit{trivial vertex invariance}.

It should be clear that both the MPP and DIV approaches 
exhibit these invariances. 
Neither method is affected by 
how the vertices are specified, so both meet the criteria for being shape-centric.

In contrast, MPP and DIV diverge on the desireable property of \textit{continutity} \cite{Mai24}.
DIV encoding is not continuous: it may contain discrete jumps 
due to infinitesimal changes in the encoded geometry, such as a slight rotation or 
position shift, if such a change causes any part of the shape 
to cross a border between tiles. 
In contrast, for MPP, there is at most an infinitesimal change 
in the encoding due to any infinitesimal change in the shape itself. 
[The qualifier \textit{at most} is necessary because some infinitestimal changes may have no affect at all
on the encoding, if the change occurs entirely within the zone of ambiguity
in Figure \ref{DecodingFig}].

\begin{figure}[b]
	\centering
	\includegraphics[width=\linewidth]{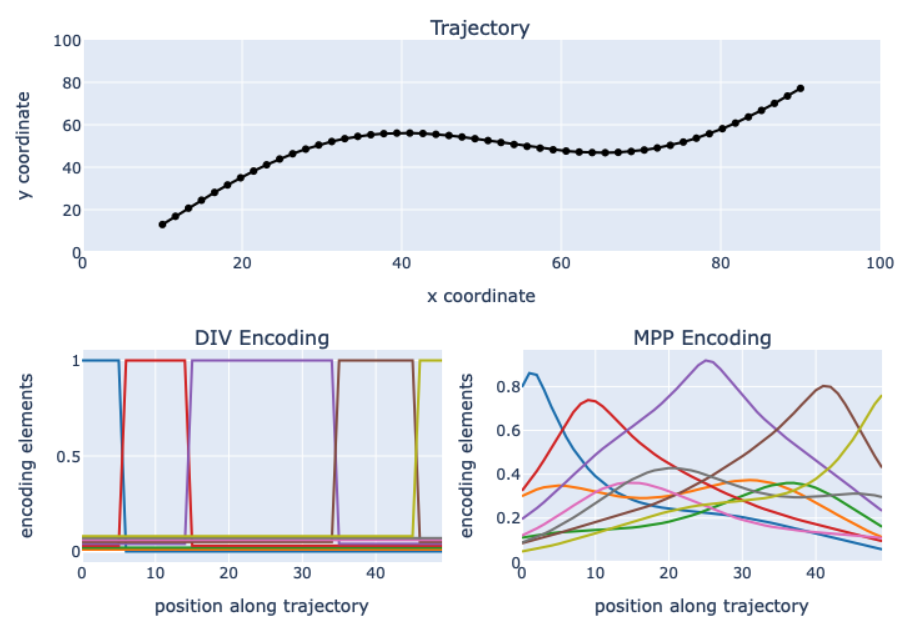}
	\caption{
		Changes in DIV and MPP encodings for a Point moving along a trajectory.
		A. Trajectory followed by the Point. 
		B. Elements of the associated DIV encoding.
		C. Elements of the associated MPP encoding.
	}
	\label{ContinuityFig}
\end{figure}

Figure \ref{ContinuityFig} illustrates this point. 
It shows changes in the values of 9-element DIV and MPP encodings for a Point
as it moves along the trajectory shown Figure \ref{ContinuityFig}A. 
Jumps in the value of the DIV encoding (Figure \ref{ContinuityFig}B) 
occur when the Point crosses into a new tile. 
In contrast, all elements of the MPP encoding (Figure \ref{ContinuityFig}C) vary continuously. 
Stated another way, for the 50 trajectory points shown here
there are only 5 unique DIV encodings. But every MPP encoding is different. 

\subsection{Estimating Geometric Properties From Encodings}

To prove that an encoding captures some geometric property,
it is sufficient to show that its value can be 
retrieved from the encoding via appropriate analysis.
In this section we present a pair of analyses to investigate that potential 
for properties of individual geometries (Section \ref{SSGP}) 
and for relationships between 
pairs of geometries (Section \ref{PairwiseSection})

\subsubsection{Single-Shape Geometric Properties}
\label{SSGP}
The spatial properties that can be computed for a geometry depend on its type.
For Points, the possible properties are few, and their representations in DIV and MPP encodings
are easily understood from discussions in the previous section.
Specifically, for DIV encodings, the location is known 
to the resolution of a grid cell,
and for MPP encodings it is known exactly. 
As there is little more that can be said about the geometric properties of a Point, 
this section will focus on LineStrings and Polygons.

\begin{figure*}
	\centering
	\includegraphics[width=\linewidth]{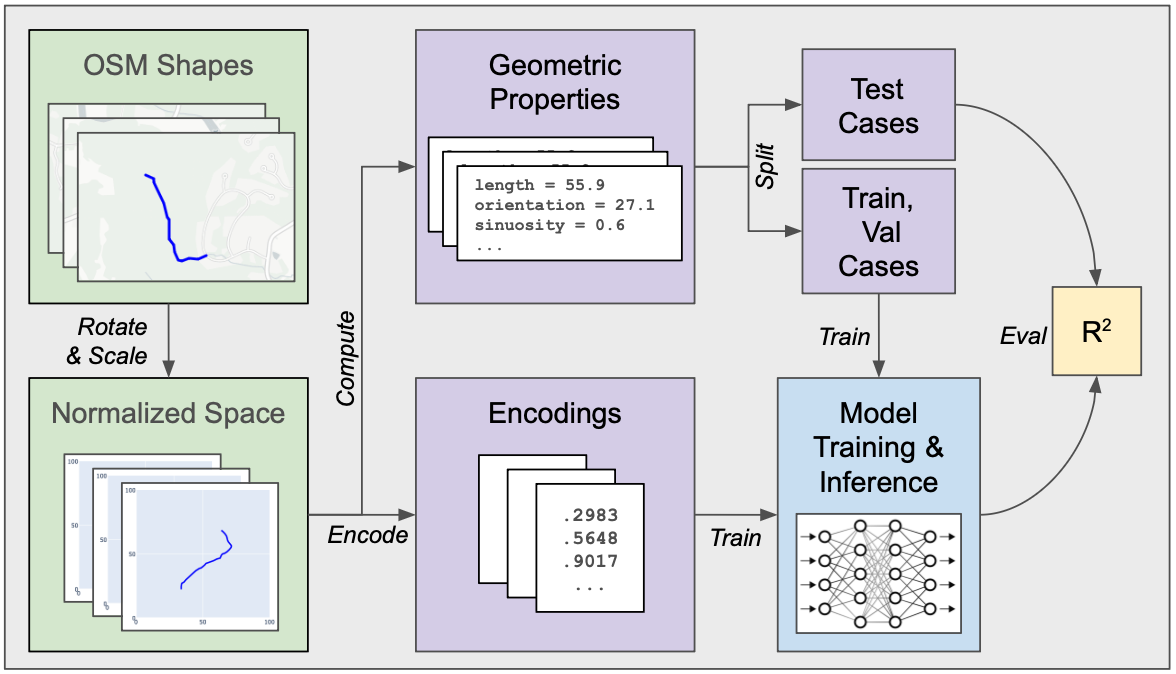}
	\caption{
		A framework for evaluating an encoding's ability to capture geometric properties.
	}
	\label{FrameworkFig}
\end{figure*}

Figure \ref{FrameworkFig} shows a framework for evaluating an encoding's ability to capture 
properties of the geometries that it represents.
The process is based on vector representations of real-world geospatail features
downloaded from OpenStreetMap (OSM).
The features are subject to random rotation and scaling, 
and are randomly placed into a normalized reference frame
with dimensions of $100 \times 100$ units.
Various geometric properties are computed directly from the vector-format representation.
The shapes are then encoded, and the encodings are used to train a neural network model
to predict the geometric properties. A suitable performance metric then quantifies the 
model's predictive power, and by implication the degree to which the measured properties
are captured by the encoding.

For this analysis we collected a set of LineStrings
by downloading major road segments and 
linear water features (rivers, streams, canals, etc.)
To get a set of Polygons, 
we downloaded town boundaries and landuse zones
(commercial districts, residential areas, forest, etc.) 
We did this for fifty 30km $\times$ 30km regions 
centered on random points around the Northeastern United States.
By applying the random transformations described above,
we created a dataset consisting of 
40,000 LineString and 40,000 Polygon geometries. 

For each shape, we computed metrics describing their fundamental properties, 
their orientation, and their complexity.
\begin{itemize}
	
\item{\textit{Fundamental properties.} For LineString objects we computed their
length, and for Polygon objects their area.}

\item{\textit{Orientation.} For both LineString and Polygon objects
	we defined an orientation by finding the two most distant 
	points in the geometry. The vector between these two points defines the 
	geometry's orientation as an angle relative to the positive x-axis. }

\item{\textit{Measures of spatial complexity.}
For LineString objects, we computed a measure of \textit{sinuosity}: 
$1 - \exp[-r / r^* + 1]$, where $r$ is the length of the LineString and $r^*$ is the 
straight-line distance between its endpoints. For a straight line segment this is zero 
and for an infinitely complex segment it is 1. 
For Polygons we computed the Convex Hull Area Ratio (CHAR):
the ratio of the polygon's area to that of its convex hull. 
Convex Polygons have a CHAR value of one
while Polygons with irregular boundaries 
have smaller values.}

\end{itemize}

The goal of this analysis is to see how well one can estimate these metrics
based on encoded DIV and MPP representations. We varied the resolution 
of the respective encoders (cell size for DIV, reference point spacing
for MPP) over the set of values shown in Table \ref{ResolutionTable}.

\begin{table}
  \caption{Encoder resolutions for testing retrieval of geometric properties.}
  \label{ResolutionTable}
  \begin{tabular}{ccc}
    \toprule
    Resolution&Grid Size&Encoding Size\\
    \midrule
    50 & $2 \times 2$  & 4 \\
    25 & $4 \times 4$  & 16 \\
    20 & $5 \times 5$  & 25 \\
    12.5 & $8 \times 8$  & 64 \\
    10 & $10 \times 10$  & 100 \\
    6.25 & $16 \times 16$  & 256 \\
    \bottomrule
  \end{tabular}
\end{table}

For every geometric property, we divided the 40,000 geometries
into training, validation, and testing sets using a 60 / 20 / 20 split.
Then we trained a neural network to predict these properties from 
the geometries' encoded representations. 
All neural networks had the same form: an input consisting of 
an encoding, two fully connected linear layers with hidden size of 128 each,
and a single output neuron. 
The models were trained on the training set, and the validation set 
was used to select the best performing model out of 3000 training epochs.
All performance metrics were computed using the 
test set.

\begin{figure*}
	\centering
	\includegraphics[width=\linewidth]{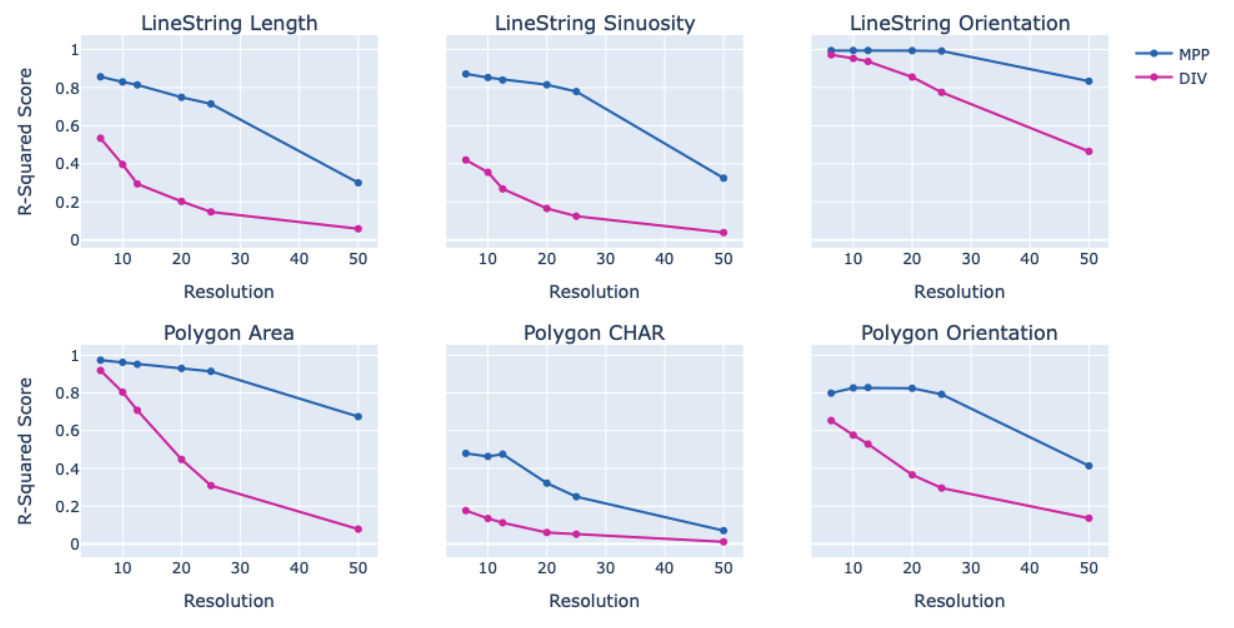}
	\caption{
		Estimation of geometric properties from MPP and DIV encodings.
	}
	\label{PropertiesFig}
\end{figure*}

As all of the properties here are continuous-valued, 
the model was trained to minimize a regression objective: mean squared error.
The performance metric used to assess the ability to estimate geometric 
properties is the coefficient of determination ($R^2$) over the test set.
The $R^2$ is the fraction of the variability in the property
that is captured by the model. In this context, this is taken to represent
the relative ability of an encoding to capture the property.

Estimating orientation required a slight modification relative to the other metrics.
Instead of estimating the orientation angle $\theta$ directly, 
we instead trained two models
to estimate $\cos(2\theta)$ and $\sin(2\theta)$. This has two benefits: 
(1) it gets around the problem of handling the discontinuity of angle values
at 360 / 0 degrees, and (2) it assures that orientation angles that differ
by 180 degrees are predicted to have the same value. An actual orientation angle
can be retrieved from these two values.  In this case the reported performance metric
is the pooled $R^2$ over the two separate predictions. 

Results of the analysis are in Figure \ref{PropertiesFig}.
In all cases MPP encoding out-performs DIV encoding.
Additionally the performance of MPP encoding is less sensitive to degrading resolution,
offering the potential to reduce the volume of data and processing required for 
any given analysis.
Overall the fidelity differs between metrics. 
Some cases yield nearly perfect retreival at finer resolutions,
including LineString orientation and Polygon area.
Other properrties are not retrieved as well, especially measures of polygon complexity.
Nontheless in all cases there is a moderate to strong positive association
between the true property values and their estimates,
indicating that the encodings capture at least some of the geometric characteristics
that may be relevant in geospatial analysis problems.

\subsubsection{Pairwise Geometric Properties}
\label{PairwiseSection}

Geospatial analysis depends on the ability to quantify relationships between geometries.
Examples include determining whether a Point lies inside or outside a Polygon,
whether a pair of LineStrings intersect, and so on.
Here we extend the analysis of the preceding section
to determine the degree to which encoded representations
can capture pairwise spatial relationships. 
The specific set of relationships that we will test are 
\begin{itemize}
	\item{\textbf{Point-In-Polygon}: whether a given Point lies in a given Polygon's interior},
	\item{\textbf{Point-On-LineString}: whether a given Point lies on a given LineString},	
	\item{\textbf{LineString-Intersects-LineString}: whether two LineStrings share at least one Point},
	\item{\textbf{LineString-Intersects-Polygon}: whether any part of a given LineString lies in the interior of a given Polygon},
	\item{\textbf{Polygon-Intersects-Polygon}: whether there is any overlap between the interiors of two Polygons},
	\item{\textbf{Polygon-Borders-Polygon}: whether two Polygons share part of thier boundaries, without sharing any part of their interiors}.
\end{itemize}
Selection and re-formatting of geometries to be used in this analysis
followed the same procedures as described in the preceding section.
For a given type of relationship, we selected a pair of geometries 
of the appropriate type 
(for example a LineString and a Polygon for the "Linestring-Intersects-Polygon" relationship)
and shifted their positions to create cases where the relationship was true
and others where it was false. 
In this manner we generated 20,000 true and 20,000 false cases
for each of the six relationship types to be tested.

The internal structure of the model was the same as for estimation of single-shape properties,
consisting of two fully connected linear layers followed by an output layer predicting a single value.
In this case the model's input is a concatenation of the encodings 
for a pair of geometries. As the relationships here are all boolean-valued,
the model was trained to minimize a binary cross-entropy loss for predicting whether
the relationship was true or false for the input shapes.

A useful perfromance metric to use in this case is the Area Under the
Receiver Operating Characteristics Curve (ROC AUC) for the prediction. 
This metric can be interrpreted as the probability that the model yields a larger prediction
for a true case than for a false one. 
 
Here again, MPP encodings outperform DIV encodings in all cases, 
and do better at maintainng good performance as resolution decreases.
ROC AUC values are above 0.95 in nearly all cases, with the exception being
the Polygon-Borders-Polygon relationship. This relationship is fairly precisely defined,
as a situation where two polygons share some portion of their borders, with no overlap of their
interiors. Cases where there is even a slight overlap or a very small space between pair of Polygons
are categorized as false, although they would tend to look very much like true cases
in an approximate encoding.

\begin{figure*}
	\centering
	\includegraphics[width=\linewidth]{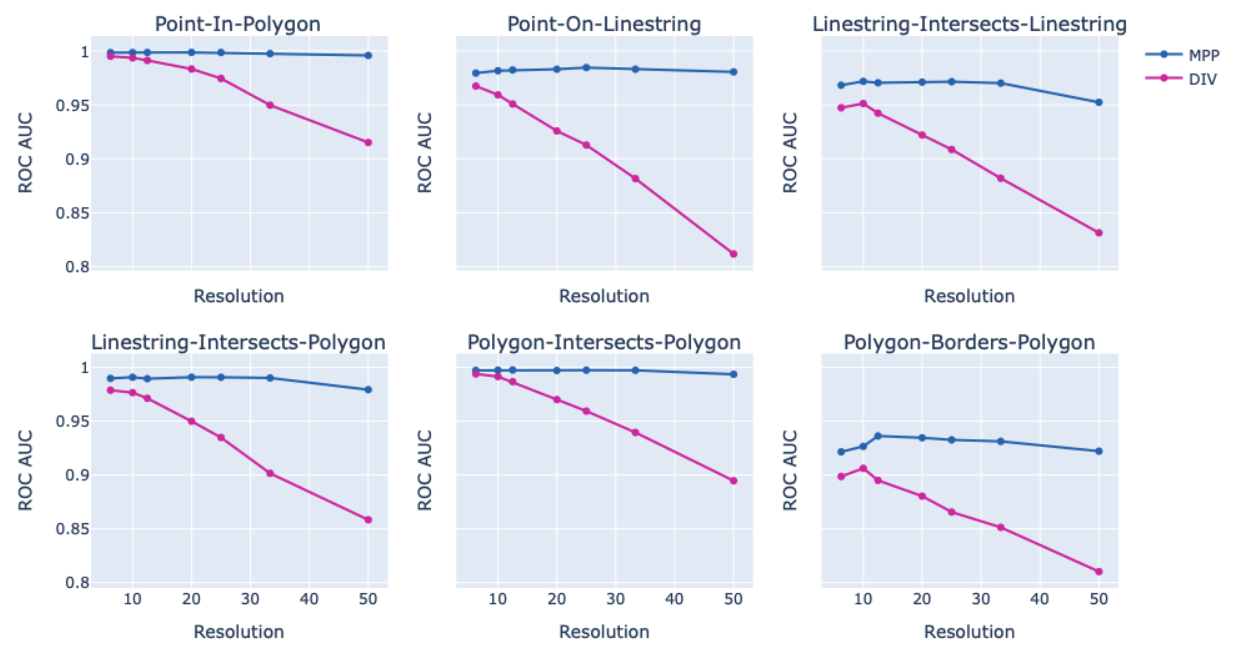}
	\caption{
		Estimation of pairwise geometric relationships from MPP and DIV encodings.
	}
	\label{PairwiseFig}
\end{figure*}

\section {Discussion}

This paper has presented
a pair of approaches for geometric encoding of arbitrary geometries
for use with Machine Learning and Artificial Intelligence models.
Analysis shows that any vector-format shape can be encoded in a way that captures 
its properties and pairwise relationships among shapes
within a given region. 
Of the two approaches examined, the one based on Multi-Point Proximity 
is clearly more capable of representing important geometric properties
than is an approach based on rasterization.

Having established the basic viability of the approach, a number of research topics remain
to be investigated.

\textit{Application to multi-part geometries}. 
All analyses in this paper have focused on the three primitive types of spatial objects:
Points, LineStrings, and Polygons. However for many analyses it is convenient to aggregate multiple
primitive shapes and to treat them as a single object, yielding geometries of type
MultiPoint, MultiLineString, Mulltipolygon, and the more general GeometryCollection \cite{ogc:sfa}.
For exmaple it may be convenient to represent all bus stops in a city 
as a single MultiPoint object rather than as a collection of individual Points.
Or all of the rivers in a watershed may be well represented as a MultiLineString object.
The MPP and DIV methods discussed here require no modificaiton to account for multi-part geometries.
The core geometric relationships defining these encodings -- tile / geometry intersection for DIV 
and closest-point distance for MPP -- are as well defined for multi-part geometries as for 
primitive types. However the increased complexity of multi-part geometries 
merits further investigation into the fidelity of their encoded representations. 

\textit{Sparse representation of MPP encodings}.
While the MPP approach out-performs DIV encodings in all analyses, the latter has one potential benefit:
DIV encodings are trivially represented as sparse vectors. 
If a large region of interest is to be encoded at a fine resolution, 
the size of the encoding vectors can become difficult to handle. 
But a DIV encoding for a relatively small object in such a case would consist mostly of zeros,
and computations on such encodings -- distance calculation in embedding space for example --
can be done using efficient sparse-vector computations. While MPP encoding as defined
in Equation \ref{MppDefEq} produces a dense vector, it easily could be made sparse by 
setting all values below a threshold to zero. Then computations on MPP encodings
could use efficient calcualtion methods as well, likely with minimal loss of representation fidelity.

\textit{Use of non-rectangular grids}.
Analyses in this paper have used rectangular grids of tiles and reference points,
covering rectangular regions.
While this is computationally simple, and corresponds closely to traditional raster-mode geospatial approaches,
there is nothing in the definition of either DIV or MPP methods that requires it.
A particular alternative is to use hexagonal grids such as the H3 reference system \cite{Sahr11}.
The centroids of all hexagons at a given H3 resolution can serve as 
the reference points for MPP encoding.
Associating an encoding model with a world-wide reference system, together with the possibility of 
sparse vector representations, offers the potential to use the MPP method
as a global system for encoding geospatial entities.

\textit{Integration with ML / AI models}.
Geometric encodings are directly ML-compatible,
and can be used as input to clustering, regression, and neural network models.
For applications that rely entirely on geometric properties, that may be sufficient.
An example application may be assessing similarity in the distributions of 
multiple types of businesses within an urban area. 
But most applications
require information on non-geometric attributes as well.
This can be handled by treating geometric encodings similarly to 
positional encodings in language models.
In the original Transformer architecture \cite{Vaswani17}, 
positional encodings are added to the encodings of text tokens, differentiating
the order in which they appear in the input. 
In the analogous geospatial setting, a model's inputs may consist of attributes of geospatial
features within an area. Any linear ordering of such inputs will fail to fully capture
their geometric relationships, which is the distinguishing feature of geospatial analysis. 
In this case, geometric encodings can serve the same role as positional text encodings:
when added to or concatenated with attribute vectors, they provide the model 
with information needed to take spatial relationships into account.

\begin{acks}
This work was performed as independent research and development 
with no external sources of funding.
\end{acks}

\bibliographystyle{ACM-Reference-Format}
\bibliography{references}


\begin{thebibliography}{28}


\ifx \showCODEN    \undefined \def \showCODEN     #1{\unskip}     \fi
\ifx \showISBNx    \undefined \def \showISBNx     #1{\unskip}     \fi
\ifx \showISBNxiii \undefined \def \showISBNxiii  #1{\unskip}     \fi
\ifx \showISSN     \undefined \def \showISSN      #1{\unskip}     \fi
\ifx \showLCCN     \undefined \def \showLCCN      #1{\unskip}     \fi
\ifx \shownote     \undefined \def \shownote      #1{#1}          \fi
\ifx \showarticletitle \undefined \def \showarticletitle #1{#1}   \fi
\ifx \showURL      \undefined \def \showURL       {\relax}        \fi
\providecommand\bibfield[2]{#2}
\providecommand\bibinfo[2]{#2}
\providecommand\natexlab[1]{#1}
\providecommand\showeprint[2][]{arXiv:#2}

\bibitem[Butler et~al\mbox{.}(2016)]%
        {rfc7946}
\bibfield{author}{\bibinfo{person}{H. Butler}, \bibinfo{person}{M. Daly},
  \bibinfo{person}{A. Doyle}, \bibinfo{person}{S. Gillies}, \bibinfo{person}{S.
  Hagen}, {and} \bibinfo{person}{T. Schaub}.} \bibinfo{year}{2016}\natexlab{}.
\newblock \bibinfo{title}{The GeoJSON Format}.
\newblock \bibinfo{howpublished}{Internet Engineering Task Force (IETF)}.
\newblock
\urldef\tempurl%
\url{https://datatracker.ietf.org/doc/html/rfc7946}
\showURL{%
\tempurl}
\newblock
\shownote{RFC 7946}.


\bibitem[Charles et~al\mbox{.}(2017)]%
        {Qi17}
\bibfield{author}{\bibinfo{person}{R.~Qi Charles}, \bibinfo{person}{Hao Su},
  \bibinfo{person}{Mo Kaichun}, {and} \bibinfo{person}{Leonidas~J. Guibas}.}
  \bibinfo{year}{2017}\natexlab{}.
\newblock \showarticletitle{PointNet: Deep Learning on Point Sets for 3D
  Classification and Segmentation}. In \bibinfo{booktitle}{\emph{2017 IEEE
  Conference on Computer Vision and Pattern Recognition (CVPR)}}.
  \bibinfo{publisher}{IEEE}, \bibinfo{address}{New York, NY, USA},
  \bibinfo{pages}{77--85}.
\newblock
\href{https://doi.org/10.1109/CVPR.2017.16}{doi:\nolinkurl{10.1109/CVPR.2017.16}}


\bibitem[Cui et~al\mbox{.}(2025)]%
        {Cui25}
\bibfield{author}{\bibinfo{person}{Longfei Cui}, \bibinfo{person}{Xinyu Niu},
  \bibinfo{person}{Haizhong Qian}, \bibinfo{person}{Xiao Wang}, {and}
  \bibinfo{person}{Junkui Xu}.} \bibinfo{year}{2025}\natexlab{}.
\newblock \showarticletitle{A Transformer-Based Approach for Efficient
  Geometric Feature Extraction from Vector Shape Data}.
\newblock \bibinfo{journal}{\emph{Applied Sciences}} \bibinfo{volume}{15},
  \bibinfo{number}{5} (\bibinfo{year}{2025}).
\newblock
\showISSN{2076-3417}
\href{https://doi.org/10.3390/app15052383}{doi:\nolinkurl{10.3390/app15052383}}


\bibitem[Gavade and Gavade(2025)]%
        {Gavade25}
\bibfield{author}{\bibinfo{person}{Anil~B. Gavade} {and}
  \bibinfo{person}{Priyanka~A. Gavade}.} \bibinfo{year}{2025}\natexlab{}.
\newblock \bibinfo{booktitle}{\emph{Explainable AI in Transforming Land Use
  Land Cover Classification}}.
\newblock \bibinfo{publisher}{Springer Nature Switzerland},
  \bibinfo{address}{Cham}, \bibinfo{pages}{343--356}.
\newblock
\showISBNx{978-3-031-75968-0}
\href{https://doi.org/10.1007/978-3-031-75968-0_18}{doi:\nolinkurl{10.1007/978-3-031-75968-0_18}}


\bibitem[Hoffer and Ailon(2015)]%
        {Hoffer15}
\bibfield{author}{\bibinfo{person}{Elad Hoffer} {and} \bibinfo{person}{Nir
  Ailon}.} \bibinfo{year}{2015}\natexlab{}.
\newblock \showarticletitle{Deep Metric Learning Using Triplet Network}. In
  \bibinfo{booktitle}{\emph{Similarity-Based Pattern Recognition}},
  \bibfield{editor}{\bibinfo{person}{Aasa Feragen}, \bibinfo{person}{Marcello
  Pelillo}, {and} \bibinfo{person}{Marco Loog}} (Eds.).
  \bibinfo{publisher}{Springer International Publishing},
  \bibinfo{address}{Cham}, \bibinfo{pages}{84--92}.
\newblock
\showISBNx{978-3-319-24261-3}


\bibitem[Jean et~al\mbox{.}(2019)]%
        {Jean19}
\bibfield{author}{\bibinfo{person}{Neal Jean}, \bibinfo{person}{Sherrie Wang},
  \bibinfo{person}{Anshul Samar}, \bibinfo{person}{George Azzari},
  \bibinfo{person}{David Lobell}, {and} \bibinfo{person}{Stefano Ermon}.}
  \bibinfo{year}{2019}\natexlab{}.
\newblock \showarticletitle{Tile2Vec: Unsupervised Representation Learning for
  Spatially Distributed Data}.
\newblock \bibinfo{journal}{\emph{Proceedings of the AAAI Conference on
  Artificial Intelligence}} \bibinfo{volume}{33}, \bibinfo{number}{01}
  (\bibinfo{date}{Jul.} \bibinfo{year}{2019}), \bibinfo{pages}{3967--3974}.
\newblock
\href{https://doi.org/10.1609/aaai.v33i01.33013967}{doi:\nolinkurl{10.1609/aaai.v33i01.33013967}}


\bibitem[Kotaridis and Lazaridou(2021)]%
        {Kotaradis21}
\bibfield{author}{\bibinfo{person}{Ioannis Kotaridis} {and}
  \bibinfo{person}{Maria Lazaridou}.} \bibinfo{year}{2021}\natexlab{}.
\newblock \showarticletitle{Remote sensing image segmentation advances: A
  meta-analysis}.
\newblock \bibinfo{journal}{\emph{ISPRS Journal of Photogrammetry and Remote
  Sensing}}  \bibinfo{volume}{173} (\bibinfo{year}{2021}),
  \bibinfo{pages}{309--322}.
\newblock
\showISSN{0924-2716}
\href{https://doi.org/10.1016/j.isprsjprs.2021.01.020}{doi:\nolinkurl{10.1016/j.isprsjprs.2021.01.020}}


\bibitem[Krizhevsky et~al\mbox{.}(2017)]%
        {Krizhevsky17}
\bibfield{author}{\bibinfo{person}{Alex Krizhevsky}, \bibinfo{person}{Ilya
  Sutskever}, {and} \bibinfo{person}{Geoffrey~E. Hinton}.}
  \bibinfo{year}{2017}\natexlab{}.
\newblock \showarticletitle{ImageNet classification with deep convolutional
  neural networks}.
\newblock \bibinfo{journal}{\emph{Commun. ACM}} \bibinfo{volume}{60},
  \bibinfo{number}{6} (\bibinfo{date}{May} \bibinfo{year}{2017}),
  \bibinfo{pages}{84–90}.
\newblock
\showISSN{0001-0782}
\href{https://doi.org/10.1145/3065386}{doi:\nolinkurl{10.1145/3065386}}


\bibitem[Ma et~al\mbox{.}(2017)]%
        {Ma17}
\bibfield{author}{\bibinfo{person}{Lei Ma}, \bibinfo{person}{Manchun Li},
  \bibinfo{person}{Xiaoxue Ma}, \bibinfo{person}{Liang Cheng},
  \bibinfo{person}{Peijun Du}, {and} \bibinfo{person}{Yongxue Liu}.}
  \bibinfo{year}{2017}\natexlab{}.
\newblock \showarticletitle{A review of supervised object-based land-cover
  image classification}.
\newblock \bibinfo{journal}{\emph{ISPRS Journal of Photogrammetry and Remote
  Sensing}}  \bibinfo{volume}{130} (\bibinfo{year}{2017}),
  \bibinfo{pages}{277--293}.
\newblock
\showISSN{0924-2716}
\href{https://doi.org/10.1016/j.isprsjprs.2017.06.001}{doi:\nolinkurl{10.1016/j.isprsjprs.2017.06.001}}


\bibitem[Mai et~al\mbox{.}(2020a)]%
        {Mai20a}
\bibfield{author}{\bibinfo{person}{Gengchen Mai}, \bibinfo{person}{Krzysztof
  Janowicz}, \bibinfo{person}{Ling Cai}, \bibinfo{person}{Rui Zhu},
  \bibinfo{person}{Blake Regalia}, \bibinfo{person}{Bo Yan},
  \bibinfo{person}{Meilin Shi}, {and} \bibinfo{person}{Ni Lao}.}
  \bibinfo{year}{2020}\natexlab{a}.
\newblock \showarticletitle{SE-KGE: A location-aware Knowledge Graph Embedding
  model for Geographic Question Answering and Spatial Semantic Lifting}.
\newblock \bibinfo{journal}{\emph{Transactions in GIS}} \bibinfo{volume}{24},
  \bibinfo{number}{3} (\bibinfo{year}{2020}), \bibinfo{pages}{623--655}.
\newblock
\href{https://doi.org/10.1111/tgis.12629}{doi:\nolinkurl{10.1111/tgis.12629}}
\showeprint{https://onlinelibrary.wiley.com/doi/pdf/10.1111/tgis.12629}


\bibitem[Mai et~al\mbox{.}(2021)]%
        {Mai22}
\bibfield{author}{\bibinfo{person}{Gengchen Mai}, \bibinfo{person}{Krzysztof
  Janowicz}, \bibinfo{person}{Yingjie Hu}, \bibinfo{person}{Song Gao},
  \bibinfo{person}{Bo Yan}, \bibinfo{person}{Rui Zhu}, \bibinfo{person}{Ling
  Cai}, {and} \bibinfo{person}{Ni Lao}.} \bibinfo{year}{2021}\natexlab{}.
\newblock \showarticletitle{A Review of Location Encoding for GeoAI: Methods
  and Applications}.
\newblock \bibinfo{journal}{\emph{International Journal of Geographical
  Information Science}}  \bibinfo{volume}{abs/2111.04006}
  (\bibinfo{year}{2021}), \bibinfo{numpages}{32}~pages.
\newblock
\showeprint[arXiv]{2111.04006}
\urldef\tempurl%
\url{https://arxiv.org/abs/2111.04006}
\showURL{%
\tempurl}


\bibitem[Mai et~al\mbox{.}(2020b)]%
        {Mai20b}
\bibfield{author}{\bibinfo{person}{Gengchen Mai}, \bibinfo{person}{Krzysztof
  Janowicz}, \bibinfo{person}{Bo Yan}, \bibinfo{person}{Rui Zhu},
  \bibinfo{person}{Ling Cai}, {and} \bibinfo{person}{Ni Lao}.}
  \bibinfo{year}{2020}\natexlab{b}.
\newblock \bibinfo{title}{Multi-Scale Representation Learning for Spatial
  Feature Distributions using Grid Cells}.
\newblock
\showeprint[arXiv]{2003.00824}
\urldef\tempurl%
\url{https://arxiv.org/abs/2003.00824}
\showURL{%
\tempurl}


\bibitem[Mai et~al\mbox{.}(2025)]%
        {Mai25}
\bibfield{author}{\bibinfo{person}{Gengchen Mai}, \bibinfo{person}{Yiqun Xie},
  \bibinfo{person}{Xiaowei Jia}, \bibinfo{person}{Ni Lao},
  \bibinfo{person}{Jinmeng Rao}, \bibinfo{person}{Qing Zhu},
  \bibinfo{person}{Zeping Liu}, \bibinfo{person}{Yao-Yi Chiang}, {and}
  \bibinfo{person}{Junfeng Jiao}.} \bibinfo{year}{2025}\natexlab{}.
\newblock \showarticletitle{Towards the next generation of Geospatial
  Artificial Intelligence}.
\newblock \bibinfo{journal}{\emph{International Journal of Applied Earth
  Observation and Geoinformation}}  \bibinfo{volume}{136}
  (\bibinfo{year}{2025}), \bibinfo{pages}{104368}.
\newblock
\showISSN{1569-8432}
\href{https://doi.org/10.1016/j.jag.2025.104368}{doi:\nolinkurl{10.1016/j.jag.2025.104368}}


\bibitem[Mai et~al\mbox{.}(2024)]%
        {Mai24}
\bibfield{author}{\bibinfo{person}{Gengchen Mai}, \bibinfo{person}{Xiaobai
  Yao}, \bibinfo{person}{Yiqun Xie}, \bibinfo{person}{Jinmeng Rao},
  \bibinfo{person}{Hao Li}, \bibinfo{person}{Qing Zhu}, \bibinfo{person}{Ziyuan
  Li}, {and} \bibinfo{person}{Ni Lao}.} \bibinfo{year}{2024}\natexlab{}.
\newblock \showarticletitle{SRL: Towards a General-Purpose Framework for
  Spatial Representation Learning}. In \bibinfo{booktitle}{\emph{Proceedings of
  the 32nd ACM International Conference on Advances in Geographic Information
  Systems}} (Atlanta, GA, USA) \emph{(\bibinfo{series}{SIGSPATIAL '24})}.
  \bibinfo{publisher}{Association for Computing Machinery},
  \bibinfo{address}{New York, NY, USA}, \bibinfo{pages}{465–468}.
\newblock
\showISBNx{9798400711077}
\href{https://doi.org/10.1145/3678717.3691246}{doi:\nolinkurl{10.1145/3678717.3691246}}


\bibitem[Mushore et~al\mbox{.}(2017)]%
        {Mushore17}
\bibfield{author}{\bibinfo{person}{Terence~Darlington Mushore},
  \bibinfo{person}{John Odindi}, \bibinfo{person}{Timothy Dube},
  \bibinfo{person}{Trylee~Nyasha Matongera}, {and} \bibinfo{person}{Onisimo
  Mutanga}.} \bibinfo{year}{2017}\natexlab{}.
\newblock \showarticletitle{Remote sensing applications in monitoring urban
  growth impacts on in-and-out door thermal conditions: A review}.
\newblock \bibinfo{journal}{\emph{Remote Sensing Applications: Society and
  Environment}}  \bibinfo{volume}{8} (\bibinfo{year}{2017}),
  \bibinfo{pages}{83--93}.
\newblock
\showISSN{2352-9385}
\href{https://doi.org/10.1016/j.rsase.2017.08.001}{doi:\nolinkurl{10.1016/j.rsase.2017.08.001}}


\bibitem[NV(nd)]%
        {Sentiance}
\bibfield{author}{\bibinfo{person}{Sentiance NV}.}
  \bibinfo{year}{n.d.}\natexlab{}.
\newblock \bibinfo{title}{Loc2Vec: Learning location embeddings with
  triplet-loss networks}.
\newblock
\urldef\tempurl%
\url{https://www.sentiance.com/2018/05/03/venue-mapping/}
\showURL{%
\tempurl}


\bibitem[{Open Geospatial Consortium}(2011a)]%
        {ogc:wkt}
\bibfield{author}{\bibinfo{person}{{Open Geospatial Consortium}}.}
  \bibinfo{year}{2011}\natexlab{a}.
\newblock \bibinfo{booktitle}{\emph{OpenGIS{\textregistered} Implementation
  Standard for Geographic information -- Simple feature access -- Part 1:
  Common architecture}}.
\newblock \bibinfo{type}{{T}echnical {R}eport} OGC 06-103r4.
  \bibinfo{institution}{Open Geospatial Consortium}.
\newblock
\urldef\tempurl%
\url{http://www.opengeospatial.org/standards/sfa}
\showURL{%
\tempurl}
\newblock
\shownote{Annex B specifies the Well-Known Text (WKT) representation of
  geometries}.


\bibitem[{Open Geospatial Consortium}(2011b)]%
        {ogc:sfa}
\bibfield{author}{\bibinfo{person}{{Open Geospatial Consortium}}.}
  \bibinfo{year}{2011}\natexlab{b}.
\newblock \bibinfo{booktitle}{\emph{OpenGIS® Implementation Standard for
  Geographic Information – Simple Feature Access – Part 1: Common
  Architecture}}.
\newblock \bibinfo{type}{{T}echnical {R}eport} OGC 06-103r4.
  \bibinfo{institution}{Open Geospatial Consortium}.
\newblock
\urldef\tempurl%
\url{http://www.opengeospatial.org/standards/sfa}
\showURL{%
\tempurl}


\bibitem[Redmon et~al\mbox{.}(2016)]%
        {Redmon16}
\bibfield{author}{\bibinfo{person}{Joseph Redmon}, \bibinfo{person}{Santosh
  Divvala}, \bibinfo{person}{Ross Girshick}, {and} \bibinfo{person}{Ali
  Farhadi}.} \bibinfo{year}{2016}\natexlab{}.
\newblock \bibinfo{title}{You Only Look Once: Unified, Real-Time Object
  Detection}.
\newblock
\showeprint[arxiv]{1506.02640}~[cs.CV]
\urldef\tempurl%
\url{https://arxiv.org/abs/1506.02640}
\showURL{%
\tempurl}


\bibitem[Sahr(2011)]%
        {Sahr11}
\bibfield{author}{\bibinfo{person}{Kevin Sahr}.}
  \bibinfo{year}{2011}\natexlab{}.
\newblock \showarticletitle{Hexagonal discrete global grid systems for
  geospatial computing}.
\newblock \bibinfo{journal}{\emph{Archives of Photogrammetry, Cartography and
  Remote Sensing}}  \bibinfo{volume}{22} (\bibinfo{date}{01}
  \bibinfo{year}{2011}).
\newblock


\bibitem[Tang et~al\mbox{.}(2015)]%
        {Tang15}
\bibfield{author}{\bibinfo{person}{Kevin Tang}, \bibinfo{person}{Manohar
  Paluri}, \bibinfo{person}{Li Fei-Fei}, \bibinfo{person}{Rob Fergus}, {and}
  \bibinfo{person}{Lubomir Bourdev}.} \bibinfo{year}{2015}\natexlab{}.
\newblock \showarticletitle{{ Improving Image Classification with Location
  Context }}. In \bibinfo{booktitle}{\emph{2015 IEEE International Conference
  on Computer Vision (ICCV)}}. \bibinfo{publisher}{IEEE Computer Society},
  \bibinfo{address}{Los Alamitos, CA, USA}, \bibinfo{pages}{1008--1016}.
\newblock
\showISSN{2380-7504}
\href{https://doi.org/10.1109/ICCV.2015.121}{doi:\nolinkurl{10.1109/ICCV.2015.121}}


\bibitem[Tobler(1970)]%
        {Tobler70}
\bibfield{author}{\bibinfo{person}{W.~R. Tobler}.}
  \bibinfo{year}{1970}\natexlab{}.
\newblock \showarticletitle{A Computer Movie Simulating Urban Growth in the
  Detroit Region}.
\newblock \bibinfo{journal}{\emph{Economic Geography}} \bibinfo{volume}{46},
  \bibinfo{number}{sup1} (\bibinfo{year}{1970}), \bibinfo{pages}{234--240}.
\newblock
\href{https://doi.org/10.2307/143141}{doi:\nolinkurl{10.2307/143141}}


\bibitem[Vaswani et~al\mbox{.}(2017)]%
        {Vaswani17}
\bibfield{author}{\bibinfo{person}{Ashish Vaswani}, \bibinfo{person}{Noam
  Shazeer}, \bibinfo{person}{Niki Parmar}, \bibinfo{person}{Jakob Uszkoreit},
  \bibinfo{person}{Llion Jones}, \bibinfo{person}{Aidan~N. Gomez},
  \bibinfo{person}{\L{}ukasz Kaiser}, {and} \bibinfo{person}{Illia
  Polosukhin}.} \bibinfo{year}{2017}\natexlab{}.
\newblock \showarticletitle{Attention is all you need}. In
  \bibinfo{booktitle}{\emph{Proceedings of the 31st International Conference on
  Neural Information Processing Systems}} (Long Beach, California, USA)
  \emph{(\bibinfo{series}{NIPS'17})}. \bibinfo{publisher}{Curran Associates
  Inc.}, \bibinfo{address}{Red Hook, NY, USA}, \bibinfo{pages}{6000–6010}.
\newblock
\showISBNx{9781510860964}


\bibitem[Wang et~al\mbox{.}(2020)]%
        {Wang20}
\bibfield{author}{\bibinfo{person}{Zhecheng Wang}, \bibinfo{person}{Haoyuan
  Li}, {and} \bibinfo{person}{Ram Rajagopal}.} \bibinfo{year}{2020}\natexlab{}.
\newblock \bibinfo{title}{Urban2Vec: Incorporating Street View Imagery and POIs
  for Multi-Modal Urban Neighborhood Embedding}.
\newblock
\showeprint[arxiv]{2001.11101}~[cs.LG]
\urldef\tempurl%
\url{https://arxiv.org/abs/2001.11101}
\showURL{%
\tempurl}


\bibitem[Woźniak and Szymański(2021)]%
        {Wozniak21}
\bibfield{author}{\bibinfo{person}{Szymon Woźniak} {and}
  \bibinfo{person}{Piotr Szymański}.} \bibinfo{year}{2021}\natexlab{}.
\newblock \showarticletitle{hex2vec: Context-Aware Embedding H3 Hexagons with
  OpenStreetMap Tags}. In \bibinfo{booktitle}{\emph{Proceedings of the 4th ACM
  SIGSPATIAL International Workshop on AI for Geographic Knowledge Discovery}}
  \emph{(\bibinfo{series}{SIGSPATIAL ’21})}. \bibinfo{publisher}{ACM},
  \bibinfo{address}{New York City, NY, USA}, \bibinfo{pages}{61–71}.
\newblock
\href{https://doi.org/10.1145/3486635.3491076}{doi:\nolinkurl{10.1145/3486635.3491076}}


\bibitem[Xu et~al\mbox{.}(2018)]%
        {Xu18}
\bibfield{author}{\bibinfo{person}{Yanyu Xu}, \bibinfo{person}{Zhixin Piao},
  {and} \bibinfo{person}{Shenghua Gao}.} \bibinfo{year}{2018}\natexlab{}.
\newblock \showarticletitle{Encoding Crowd Interaction with Deep Neural Network
  for Pedestrian Trajectory Prediction}. In \bibinfo{booktitle}{\emph{2018
  IEEE/CVF Conference on Computer Vision and Pattern Recognition}}.
  \bibinfo{publisher}{IEEE}, \bibinfo{address}{Salt Lake City, UT, USA},
  \bibinfo{pages}{5275--5284}.
\newblock
\href{https://doi.org/10.1109/CVPR.2018.00553}{doi:\nolinkurl{10.1109/CVPR.2018.00553}}


\bibitem[Yin et~al\mbox{.}(2019)]%
        {Yin19}
\bibfield{author}{\bibinfo{person}{Yifang Yin}, \bibinfo{person}{Zhenguang
  Liu}, \bibinfo{person}{Ying Zhang}, \bibinfo{person}{Sheng Wang},
  \bibinfo{person}{Rajiv~Ratn Shah}, {and} \bibinfo{person}{Roger Zimmermann}.}
  \bibinfo{year}{2019}\natexlab{}.
\newblock \showarticletitle{GPS2Vec: Towards Generating Worldwide GPS
  Embeddings}. In \bibinfo{booktitle}{\emph{Proceedings of the 27th ACM
  SIGSPATIAL International Conference on Advances in Geographic Information
  Systems}} (Chicago, IL, USA) \emph{(\bibinfo{series}{SIGSPATIAL '19})}.
  \bibinfo{publisher}{Association for Computing Machinery},
  \bibinfo{address}{New York, NY, USA}, \bibinfo{pages}{416–419}.
\newblock
\showISBNx{9781450369091}
\href{https://doi.org/10.1145/3347146.3359067}{doi:\nolinkurl{10.1145/3347146.3359067}}


\bibitem[Zhang et~al\mbox{.}(2023)]%
        {Zhang23}
\bibfield{author}{\bibinfo{person}{Xiangrong Zhang}, \bibinfo{person}{Tianyang
  Zhang}, \bibinfo{person}{Guanchun Wang}, \bibinfo{person}{Peng Zhu},
  \bibinfo{person}{Xu Tang}, \bibinfo{person}{Xiuping Jia}, {and}
  \bibinfo{person}{Licheng Jiao}.} \bibinfo{year}{2023}\natexlab{}.
\newblock \bibinfo{title}{Remote Sensing Object Detection Meets Deep Learning:
  A Meta-review of Challenges and Advances}.
\newblock
\showeprint[arxiv]{2309.06751}~[cs.CV]
\urldef\tempurl%
\url{https://arxiv.org/abs/2309.06751}
\showURL{%
\tempurl}


\end{thebibliography}

\end{document}